\def\year{2022}\relax
\documentclass[letterpaper]{article} 
\usepackage{aaai22}  
\usepackage{times}  
\usepackage{helvet}  
\usepackage{courier}  
\usepackage[hyphens]{url}  
\usepackage{graphicx} 
\usepackage{natbib}  
\usepackage{caption} 
\DeclareCaptionStyle{ruled}{labelfont=normalfont,labelsep=colon,strut=off} 
\usepackage{array}
\usepackage{comment}
\usepackage{subfigure}
\usepackage{xcolor,colortbl} 
\usepackage{multirow}
\usepackage{makecell}
\usepackage{xspace}
\usepackage{wrapfig,lipsum,booktabs}
\usepackage{amsmath}
\usepackage{amssymb}
\usepackage{amsfonts}
\usepackage[ruled,vlined]{algorithm2e}
\usepackage{cleveref}
\usepackage{soul}
\Crefformat{equation}{(#2#1#3)}
\makeatletter
\newcommand{\printfnsymbol}[1]{%
  \textsuperscript{\@fnsymbol{#1}}%
}
\DeclareMathOperator*{\argmax}{argmax}
\newcommand{\oreo}{\textsc{oreo}\xspace}
\newcommand{\robert}{RoBERTa\xspace}
\newcommand{\newturk}{\textsc{Newsela-turk}\xspace}
\newcommand{\masker}{\textsc{Masker}\xspace}

\title{Text Revision by On-the-Fly Representation Optimization}
\author{
    Jingjing Li\textsuperscript{\rm 1},
    ~~Zichao Li\textsuperscript{\rm 2},
    ~~Tao Ge\textsuperscript{\rm 3},
    ~~Irwin King\textsuperscript{\rm 1},
    ~~Michael R. Lyu\textsuperscript{\rm 1}
}
\affiliations{
    \textsuperscript{\rm 1}The Chinese University of Hong Kong,~
    \textsuperscript{\rm 2}Mila/McGill University, \\
    \textsuperscript{\rm 3}Microsoft Research Asia \\
    
    \texttt{llee.jingjing@gmail.com} \quad \texttt{zichao.li@mail.mcgill.ca}\\
    \texttt{tage@microsoft.com} \quad \texttt{\{king, lyu\}@cse.cuhk.edu.hk}\\
    
}

\begin{document}

\maketitle

\begin{abstract}
Text revision refers to a family of natural language generation tasks, where the source and target sequences share moderate resemblance in surface form but differentiate in attributes, such as text formality and simplicity. Current state-of-the-art methods formulate these tasks as sequence-to-sequence learning problems, which rely on large-scale parallel training corpus. In this paper, we present an iterative in-place editing approach for text revision, which requires no parallel data. In this approach, we simply fine-tune a pre-trained Transformer with masked language modeling and attribute classification. During inference, the editing at each iteration is realized by two-step span replacement. At the first step, the distributed representation of the text optimizes on the fly towards an attribute function. At the second step, a text span is masked and another new one is proposed conditioned on the optimized representation. The empirical experiments on two typical and important text revision tasks, text formalization and text simplification, show the effectiveness of our approach. It achieves competitive and even better performance than state-of-the-art supervised methods on text simplification, and gains better performance than strong unsupervised methods on text formalization
\footnote{Code and model are available at \url{https://github.com/jingjingli01/OREO}}.

\end{abstract}

\section{Introduction}
Text revision refers to an important series of text generation tasks, including but not limited to text style transfer~\cite{shen2017style}, text simplification~\cite{xu2016optimizing}, counterfactual debiasing~\cite{zmigrod2019counterfactual}, grammar error correction~\cite{sun2022unified}, sentence fusion~\cite{malmi2019lasertagger} and argument reframing~\cite{chakrabarty2021argreframe}, which revises an input sentence into another one with the desired attribute (e.g., formality or simplicity). As the most popular solution, sequence-to-sequence (seq2seq) learning achieves state-of-the-art results on many text revision tasks today. However, it becomes less applicable when there is no large-scale annotated parallel data for training. 

On the other hand, recent breakthroughs in self-supervised learning have enabled the pre-trained Transformer models~\cite{vaswani2017transformer}, such as BERT~\cite{devlin2018bert}, RoBERTa~\cite{liu2019roberta} and GPT~\cite{radford2018igpt}, to learn sufficient distributed representation of natural language, which is universally transferable to a wide range of downstream tasks even without labeled data~\cite{tenney2019bertpipe, Zhang2019BERTScoreET, wu2020perturbed}. 
In this paper, we ask the question, can we borrow the power of a pre-trained Transformer for text revision without any parallel data?

There exist some efforts on developing unsupervised text generation methods with only non-parallel data, such as using reinforcement learning (RL)~\cite{yu2017seqgan} and variational auto-encoders~\cite{hu2017toward}. However, these methods suffer from issues of unstable~\cite{bowman2016vae} and computationally expensive training. It is even more challenging to apply them with large pre-trained models. For instance, to fine-tune a GPT-3 summarization model with RL, it takes thousands of labeler hours for learning a reliable reward function and 320 GPU-days to train the policy and value nets~\cite{stiennon2020openaisum}. 

\begin{figure*}[t]
    \centering
    \includegraphics[width=\textwidth]{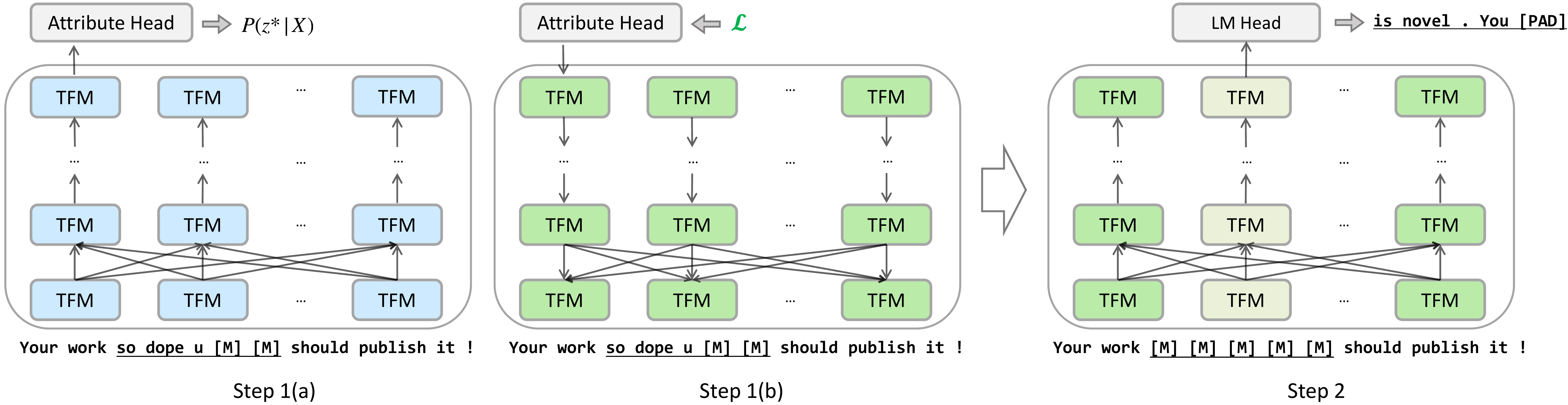}
    \caption{\small A simplified illustration of two-step span revision in \oreo. In this example, the input is ``\textit{Your work so dope u should publish it!}''. The informal textual span ``\textit{so dope u}'' is selected to revise. To allow for a potentially longer replacement, we append 2 \texttt{[LM-MASK]} tokens to the span and use this sequence for two-step revision.
Step 1: Representation Optimization. (a) The fine-tuned RoBERTa model encodes an input sentence to calculate the likelihood of target attribute $P_\theta(z^*|X)$. (b) After calculating and backpropagating the loss between estimated and target attribute value, the hidden states (in green) are optimized on the fly. Step 2: Span replacement. The span to be edited is replaced with \texttt{[LM-MASK]} tokens (we use \texttt{[M]} for short). We fix the optimized hidden representations in Step 1 (in green) and let \robert's LM head propose an alternative text span autoregressively.}
    \label{fig:text-revise}
\end{figure*}

In this work, we propose \oreo, a method of \textsc{o}n-the-fly \textsc{re}presentation \textsc{o}ptimization for text revision. Instead of generating an entire sequence of tokens from scratch, \oreo first detects partial text span to be edited, then conducts in-place span revision, which is realized by iterative mask-and-infill editing on the input sentence. As shown in \Cref{fig:text-revise}, at each iteration, a fine-tuned \robert encodes the input sentence into a distributed representation, then optimizes it informed by an attribute head of the same pretrained \robert model. After that, \oreo masks a span and infills a new one conditioned on the updated representation. As for the training, our model, \oreo fine-tunes \robert with two simple tasks, masked language modeling and attribute classification.

The contribution of this work is three-fold: 
\begin{enumerate}
    \item We propose an efficient mask-and-infill method with on-the-fly optimized representation for text revision. In this work, we tackle two important tasks: text simplification and text formalization. Additionally, this framework can be directly adapted to other text revision tasks.
    \item To enable on-the-fly representation optimization, we design simple fine-tuning methods that balance efficiency and efficacy. The fine-tuning can be finished within 8 GPU-hours at most in our experiments.
    \item Our proposed \oreo has strong performance on text formalization dataset GYAFC-fr~\cite{Rao2018DearSO}, surpassing unsupervised baseline methods, one of which also utilizes \robert; and achieves competitive performance with state-of-the-art supervised methods on text simplification dataset \newturk~\cite{maddela2020controllable}.
\end{enumerate}

\section{Methods}

\subsection{Problem Formulation}
Text revision aims to revise an input sentence $X$ with attribute $z$ to another one $X^{*}$ with the target attribute $z^{*}$, while other features fixed as much as possible. In this work, we address text simplification and text formalization, where the target attributes are simplicity and formality respectively. The training data is a non-parallel corpus with attribute labels. 

\subsection{Preliminary: Pre-trained Transformer Models for Natural Language}

Self-supervised learning with massive unlabeled text data makes powerful pre-trained Transformers for natural language processing. We adopt the RoBERTa$_{\text{base}}$~\cite{liu2019roberta} model in this work.

\robert is a stack of $L$ Transformer layers trained with masked language modeling with unlabeled text data. Given a sequence of tokens $[x_1, ..., x_T]$ with length $T$ that is partially masked (e.g. $x_t$ is replaced by a special \texttt{[MASK]} token), \robert constructs hidden states $H^l_t$ at $l$-th layer for token $x_t$. On top of the Transformer layers of \robert, there is a language model (LM) head that takes as input the hidden states $H^L_t$ at the final layer corresponding to the masked token, and recovers the masked token $x_t$ by maximizing:
\begin{equation}
    P_{W_{\text{LM}}}(x_{t}|H_{t}^L) = \text{Softmax}(W_{\text{LM}}^T H_{t}^L),
\end{equation}
where $W_{\text{LM}}$ is the parameter of LM head and $H_{\setminus t}$ is hidden states at positions other than $t$. $H_{t}$ has intensive interaction with $H_{\setminus t}$ through self-attention module. Therefore, \robert is able to infill context-aware tokens.

\subsection{Training for \oreo: Multi-task Fine-tuning}
The hidden states produced by RoBERTa, or in general, pre-trained Transformer models, have been proven to encode a wide range of linguistic features, such as morphology~\cite{li2019specializing}, syntax~\cite{wu2020perturbed}, semantics~\cite{Zhang2019BERTScoreET} and etc. Motivated by this, we fine-tune the \robert to model the task-specific attributes. Concretely, we adopt two fine-tuning tasks, masked language modeling (MLM) and attribute classification. The former one is to force \robert to infill a span consistent with the semantics and attributes encoded in the hidden states, and the latter one is to help \robert update the hidden states towards a specific attribute.

\subsubsection{Masked language modeling}
The original MLM objective adopted by \robert does not model the length of tokens to be infilled. Inspired by~\citet{malmi2020padlm}, we let the model do variant-length span replacement. Specifically, there are three modifications for the MLM objective: 1) We introduce a new special token $\texttt{[LM-MASK]}$ for span infilling; 2) Before each iteration of span replacement, we append $K$ additional masks to pad the selected span to a fixed length; 3) \robert can predict \texttt{[PAD]}, another new special token, as a placeholder to be removed directly from the output text. As such, a selected span of length $N$ can be replaced by a new one, whose length is between 0 and $N\!+\!K$. 

We modify the strategy for MLM training data construction accordingly. A continuous span is masked, and we randomly insert \texttt{[LM-MASK]} and \texttt{[PAD]} tokens in the source and target spans, respectively. We provide an example and more details in Appendix A.

Meanwhile, we still follow the original masking strategy, where tokens are masked independently and replaced by \texttt{[MASK]} token, creating another set of MLM training data. We fine-tune \robert and its LM head with two sets of training data jointly.

\subsubsection{Attribute classification}
In addition, we create a new attribute head, parallel to the LM head, on top of \robert as an attribute classifier. The conventional fine-tuning approach takes as input the outputs of the final layer at position $t=0$. In our preliminary experiment, we find this approach sub-optimal. Inspired by the evidence found in~\cite{tenney2019bertpipe} that the different layers of pre-trained Transformer capture different categories of features, we concatenate the hidden states of the \texttt{[CLS]} token from all layers as the input of attribute head. Specifically, given an input sentence $X$, \robert with parameters $\theta$ predicts the probability distribution over attribute candidates $Z$ as:
\begin{equation}
    P_{\theta}(Z|X) = \text{Softmax}(W_{\text{Att}}^T[H_0^0, H_0^1, ..., H_0^L])
\end{equation}
where $W_{\text{Att}}$ denotes parameters of the attribute head, and $[H_0^0, H_0^1, ..., H_0^L]$ is the concatenation of hidden states from all layers at the position $t=0$. Then the \robert is tuned to maximize the likelihood of ground-truth attribute labels.

\begin{algorithm}[t!]
\small
 \textbf{Input:} An input sentence $ \mathrm X^{(0)}$; \\
  \quad\quad\quad Set target attribute $z^*$, threshold $\delta$, maximum iteration number $I$;\\
 \quad\quad\quad A fine-tuned \robert with parameters $\theta$, including an attribute head  $W_{\text{Att}}$ and a LM head  $W_{\text{LM}}$\\
 
 \textbf{Output:} An output sentence $\mathrm X^*$\\

\textbf{Initialize:} $i=0$, $\zeta^{(0)}=P_\theta(z^*|X^{(0)})$\\
 \While{$i < I \operatorname{and} \zeta^{(i)} <\delta$}{
  {\color{blue}$\rhd$ Span selection} \\
 Calculate $\zeta^{(i)}=P_\theta(z^*|X^{(i)})$ and $\mathcal{L}$ \Cref{eqn:loss} \\
 Calculate $a^{(i)}$ \Cref{eqn:word-dis} and select $t, N = \argmax\limits_{t, N} a^{(i)}_{t:t+N}$ \\
  {\color{blue}$\rhd$ Representation optimization} \\
 Insert $K$ \texttt{[LM-MASK]}s after $X^{(i)}_{t:t+N}$, then we have $X'^{(i)}$ as the input of \robert at the next step\\ 
 Calculate $H^{(i)}$, $P_{W_{\text{Att}}}(z^*|H^{(i)})$ and $\mathcal{L}'$ \Cref{eqn:loss}\\
 Update $H^{(i+1)}$ with $\nabla_{H^{(i)}} \mathcal{L}'$ \Cref{eqn:update}\\
  {\color{blue}$\rhd$ Span replacement} \\
 Replace the selected span $X'^{(i)}_{t:t+N}$ with \texttt{[LM-MASK]}s\\
 $X^{(i+1)}_{\setminus t:t+N+K}=X'^{(i)}_{\setminus t:t+N+K}$\\
 \quad\quad{\color{gray}$\rhd$ The unselected part keep fixed}\\
 Infill a new span $\!X^{(i+1)}_{t\!:t\!+\!N\!+\!K}\!=\!\argmax\limits_{X_{t:t\!+\!N\!+\!K}}\!P_{W_{\text{LM}}}(X_{t\!:t\!+\!N\!+\!K}\!|H^{(i+1)}_{\setminus t:t\!+\!N\!+\!K}\!)$ \\ 
 
 \quad\quad{\color{gray}$\rhd$ Approximate by greedy decoding}\\
 Remove the \texttt{[PAD]} tokens in the new span, then we have $X^{(i+1)}$\\
 }
 \textbf{Return:} $X^*=X^{(j)}$, where $j=\argmax\limits_{j}\zeta^{(j)}$
 \caption{Text revision with \oreo}
 \label{alg:revise}
\end{algorithm}

\subsection{Inference: On-the-fly Representation Optimization}
Most of the existing work on unsupervised text generation incorporate task-specific constraints, such as reconstruction objective and discriminator networks~\cite{surya2018unts}, on the generation model explicitly. In contrast, we steer the distributed representation of text directly. The hypothesis is that the pre-training and fine-tuning make \robert an intrinsic multi-task model, which has already learned sufficient features for text revision: the hidden states can be used to recognize the attribute, and meanwhile inform the LM head to select tokens consistent to a certain attribute and context. All we need further is to keep other attributes, especially the semantics, fixed as much as possible during modification.

To this end, \oreo conducts text revision by iteratively replacing spans on the input sequence. At each iteration, a span is selected for editing; then the revision is done in two steps. At the first step, \robert encodes the input sentence into hidden states, conditioned on which the attribute head measures the probability of target attributes. Then \robert adjusts the hidden states towards increasing the target attribute probability. At the second step, the selected span is masked out, after which \robert uses the LM head to fill in the blank, conditioned on updated hidden states. 
These two steps repeatedly iterate until a maximum iteration number $I$ is reached, or the attribute value exceeds a predefined threshold $\delta$. The complete revision procedure of \oreo is formalized in \Cref{alg:revise}. 

In the following sections, we detail two steps of text revision in \oreo respectively. An illustration is provided in \Cref{fig:text-revise}. Then we introduce our method of span selection.

\subsubsection{Step 1: Representation optimization}
Given an input sentence $X^{(i)}$ at the $i$-th iteration, \robert parameterized by $\theta$ transforms it to a sequence of hidden states $H^{(i)}$, conditioned on which the attribute head estimates the probability of target attribute $P_{W_{\text{Att}}}(z^*|H^{(i)})$. However, blindly finding a $H^*$ that optimizes $P_{W_{\text{Att}}}(z^*|H^*)$ can corrupt or even eliminate other useful features encoded in the original hidden states, and we may not want those features to be greatly influenced. Thus, for each revision, we find a small local perturbation on $H^{(i)}$ that 
maximally increases the likelihood of target attribute. As such, the update rule of hidden states is:
\begin{equation}\label{eqn:update}
    \begin{aligned}
        H^{(i+1)} = H^{(i)} - \lambda 
        \frac{\nabla_{H^{(i)}}\mathcal{L}}
        {\Vert\nabla_{H^{(i)}}\mathcal{L}\Vert_2}
    \end{aligned},
\end{equation}
where $\lambda$ is a hyper-parameter that controls the norm of perturbation, and 
\begin{equation}\label{eqn:loss}
    \mathcal{L} = - \log P_{W_{\text{Att}}}(z^*|H^{(i)}).
\end{equation}
The perturbation, also known as the normalized gradient of $\mathcal{L}$ with respect to hidden states, can be calculated with standard backpropagation techniques. The parameters of \robert is frozen during this gradient computation. Therefore, the representation is optimized on-the-fly.

Even though we apply a small perturbation, there are still risks that other coupled attributes change accordingly. We address this issue by only replacing one span at each iteration, and encoding the complete sentence into hidden states before masking a span. This issue can be further eliminated by other advanced techniques, such as representation disentanglement~\cite{chen2019disentangle} and neural adapter modules~\cite{madotto2020adapter}. We leave the exploration of more advanced solutions for future work.

\subsubsection{Step 2: Span replacement}
Once the hidden states are updated, \textsc{oreo} conducts span replacement. The selected span $X_{t:t+N}^{(i)}$ of length $N$ is replaced by \texttt{[LM-MASK]} tokens. And hence the span to be infilled is $X_{t:t+N+K}^{(i)}$ (we append $K$ \texttt{[LM-MASK]} tokens before updating hidden states). \robert takes as input the masked sequence, and predicts a new span autoregressively with the previously updated hidden states:
\begin{equation}\label{eqn:decode_span}
    \begin{aligned}
    P_{W_{\text{LM}}}&(X_{t:t+N+K}^{(i+1)}|H^{(i+1)}_{\setminus t:t+N+K})=& \\ 
    \prod_{n=1}^{N+K} &P_{W_{\text{LM}}}(x_{t+n}^{(i+1)}|H^{(i+1)}_{\setminus t:t+N+K},X_{t:t+n}^{(i+1)}),
    \end{aligned}
\end{equation}
where $x_{t+n}^{(i+1)}$ is the predicted token at step $n$, $H^{(i+1)}_{\setminus t:t+N+K}$ is the optimized hidden states of unselected text. Informed by the updated hidden states, the revised span is expected to meet target attribute and meanwhile maintain other information, e.g. semantics, of the original span. 

\subsubsection{Span selection strategy}\label{sec:span-select}
The span selection in \oreo is done before the text revision at each iteration. It is motivated by three reasons: 1) The selection strategy can be agnostic to the text revision algorithm, increasing the flexibility of \oreo; 2) It allows us to insert $\texttt{[LM-MASK]}$ tokens in the selected span in advance, so that \robert can infill a longer span. 3) It enables human-in-the-loop generation, where the user can indicate which part should be revised.

In this work, we use the magnitude of the $\nabla_{H^{(i)}}\mathcal{L}$, where $\mathcal{L}$ is calculated with \Cref{eqn:loss}, as a measurement of disagreement for span selection. Specifically, at iteration $i$, we calculate $a_{t}^{(i)}$ for each token with respect to the attribute head as:
\begin{equation}\label{eqn:word-dis}
    a_t^{(i)} = \Vert\nabla_{H_t^{0^{(i)}}}\mathcal{L}\Vert_2,
\end{equation}
where $H^0$ is the hidden states at the word embedding layer. Intuitively, a token whose modification can maximally increase the target attribute value should be revised. 

Then we calculate an N-gram ($n \leq 4$) score as:
\begin{equation}\label{eqn:span-dis}
    a_{t:t+N}^{(i)} = \frac{ \sum_{n=1}^{N} a_{t+n}^{(i)} }{ N + c},
\end{equation} 
where we add a smoothing constant $c$, otherwise only one token is chosen. In practice, we set $c$ as 1. To further prevent serious corruption of the original sentence, we remove named entities from the selected span.
As mentioned above, we finally append $K$ \texttt{[LM-MASK]} tokens to the selected span for the two-step span replacement.

\section{Experiment Setting}

\subsection{Implementation}\label{subsec:implement}
We experiment with \oreo in two real-world text revision tasks, text simplification and text formalization. We implement \robert based on Huggingface transformers \cite{wolf-etal-2020-transformers}. For all experiments, we fine-tune the \robert$_{base}$~\cite{liu2019roberta} with a task-specific corpus. 
We primarily adopted the default hyperparameters with a fixed learning rate of 5e-5. The numbers of fine-tuning epochs are 6 and 2 for text simplification and formalization, respectively. It takes 8-GPU hours to fine-tune \robert on one Tesla V100 for both tasks. 
The maximum iteration $I$  was set to 4 for efficiency purpose, although the final performance can increase slightly with more iterations.
$\lambda$ was selected from $\{0.8, 1.2, 1.6, 2.0\}$ and set to 1.6. 
These parameters are validated only on the text formalization. We do not perform further tuning on text simplification. 
The attribute threshold $\delta$ is task-dependent. It was selected from from $\{0.1, 0.2, \dots ,0.5\}$ and set to 0.5 for text simplification and 0.3 for text formalization. $K=$ 1 for both tasks.

\subsection{Text Simplification}\label{subset:simp_task}

Text simplification is to revise the complex text into simpler language with easy grammar and word choice while keeping the meaning unchanged \cite{saggion2017automatic}. 
Based on the widely used corpora Newsela \cite{xu2015problems}, \citet{jiang2020neural} constructs a reliable corpus consisting of 666K complex-simple sentence pairs\footnote{Dataset available at https://github.com/chaojiang06/wiki-auto. Newsela dataset can be requested from https://newsela.com/data/}. As our model does not rely on the complex-simple alignments, we remove the duplicated sentences. The final dataset consists of 269K train, 28K development and 29K test sentences. 
As discussed in \cite{jiang2020neural, maddela2020controllable, alva2017learning}, previous supervised methods tend to behave conservatively by simply deleting words and lack the ability to conduct effective phrasal simplification, we follow \cite{maddela2020controllable} and adopt \newturk for evaluation, a test set with high-quality human-written references emphasizing lexical and phrasal simplification for each complex sentence. Although it is challenging for \oreo to conduct structural simplification, there is an off-the-shelf resource~\cite{niklaus2019dissim} focused on sentence splitting and deletion that we can utilize as a pre-processing of complex sentences. To keep this work focused, we leave structural 
transformation for future work.

We report SARI \cite{xu2016optimizing}, Flesch-Kincaid grade level (FKGL) readability \cite{kincaid1975derivation}  and average sentence length (SLen) as evaluation metrics. SARI calculates the average of F1/precision of $n$-grams added, kept and deleted between system output and reference sentences ($n \in$ \{1, 2, 3, 4\}). We report the F1 score of each edit operation. FKGL measures the readability of sentences. We do not report BLEU because it does not correlate well with human judgement~\cite{xu2016optimizing}.

\begin{table}[t!]
        \centering
        \resizebox{\linewidth}{!}{
                \begin{tabular}{l|c|ccc|c|c}
        		\toprule
        		Methods & SARI  & Add & Keep & Delete & FKGL$^\downarrow$ & SLen \\ \midrule
        		\multicolumn{7}{c}{Supervised}\\ \midrule
        		Complex (input) 
        		& 22.3 & 0.0 & 67.0 & 0.0 & 12.8 & 23.2 \\\midrule
        		Transformer$_{\text{BERT}}$ 
        		& 36.0 & 3.3 & 54.9 & 49.8 & \textbf{8.9} & 16.1 \\
        		EditNTS 
        		& 37.4 & 1.6 & 61.0 & 49.6 & 9.5 & 16.9 \\
        		Hybird-NG 
        		& 38.2 & 2.8 & 57.0 & 54.8 & 10.7 & 21.6 \\
        		ControlTextSimp 
        		& 41.0 & \textbf{3.4} & 63.1 & 56.6 & 11.5 & 22.2 \\ \midrule
        		\multicolumn{7}{c}{Unsupervised}\\ \midrule
        		UNTS 
        		& 39.9 & 1.5 & 60.5 & 57.7 & 11.2 & 22.0 \\ 
        		\oreo (ours)
        		& \textbf{45.2} & 2.3 & \textbf{69.4} & \textbf{64.0} & 11.4 & \textbf{23.5} \\ 
        		\bottomrule
        	\end{tabular}
        }
        \caption{Automatic evaluation results on \newturk. $^\downarrow$The smaller, the better.}
        \label{tab:simp_full}
\end{table}

\begin{table}[htp]
        \centering
        \resizebox{\linewidth}{!}{
            \begin{tabular}{l|c|c|cc}
        		\toprule
        		Methods$^\dag$ & BLEU  & Formality & H-mean & G-mean \\ \midrule
        		Human reference
        		& 100.0 & 95.20 & 97.49 & 97.52\\ \midrule
        		CrossAlign 
        		& 4.77 & 75.9 &  8.98 & 19.03 \\
        		StyleEmbded 
        		& 8.71 & 28.3 & 13.32 & 15.70\\ 
        		MultiDec 
        		& 14.04 & 21.32 & 16.93 & 17.30 \\ 
        		UnsupMT 
        		& 37.36 & 76.88 & 50.28 & 53.59 \\ 
        		\masker 
        		& 47.73 & 58.86 & 52.71 & 53.00 \\ \midrule
        		\oreo (ours) 
        		& \textbf{57.63} & \textbf{80.71} & \textbf{67.24} & \textbf{68.20} \\  
        		\bottomrule
    	    \end{tabular}
        }
        \caption{Automatic evaluation results on text formalization.}
        \label{tab:style_full}
\end{table}

We compare our \oreo to both supervised and unsupervised approaches. For unsupervised baselines, we adopt UNTS~\cite{surya2018unts}, which is based on adversarial training and variational auto-encoder. We also compare our model with the following state-of-the-art supervised methods:
(i) Transformer$_{\text{BERT}}$~\cite{rothe2020leveraging}, a Transformer whose encoder is initialized with the BERT model.
(ii) EditNTS~\cite{dong2019editnts}, which models edit operations explicitly with sequence-to-sequence learning. 
(iii) Hybrid-NG~\cite{narayan2014hybrid}, a hybrid system including a probabilistic model for splitting and deletion, and a monolingual machine translation model for phrase replacement and reordering.
(iv) ControlTextSimp~\cite{maddela2020controllable}, the current state-of-the-art method composed of structural simplification module and lexical/phrasal simplification model. 
We also report the performance of the strategy that blindly copies the original complex sentence.

\subsection{Text Formalization}\label{subset:style_task}

We then move on to the next task, text formalization. Since the informal sentence is much noisier than the pre-training data of \robert, this task can test the robustness of our \oreo. To compare with previous work, we experimented with the domain of Family \& Relationships in Grammarly's Yahoo Answers Formality Corpus (GYAFC-fr)~\cite{Rao2018DearSO}. 
There are 100K, 5K and 2.5K informal-formal\footnote{The informal text in GYAFC is collected from casual chats in web forums. It includes few offensive statements, such as slang, vulgarity, harassment, etc. These statements may cause discomfort or upset to the user of the dataset.} pairs in GYAFC. Again, we only use non-parallel sentences and their associated formality labels to fine-tune \robert. Considering the gap between informal text and pre-training corpus, we augment the training data with 880K automatically extracted sentences from the same domain by \citet{Xu2019FormalityST}.

The evaluation of formalization involves multiple aspects. Following previous literature~\cite{Luo19DualRL,xu2018unpaired}, we report BLEU~\cite{papineni2002bleu} as the measurement of content preservation and fluency. The formality attribute is evaluated by a separately trained \robert classifier which obtains accuracy at 94\% on the validation set. To obtain an overall performance of the system, we calculate the harmonic mean (H-mean) and geometric mean (G-mean) of BLEU and formality accuracy and consider them as the main metric for this task.

We compare \oreo with the following widely adopted unsupervised baseline methods: (i) CrossAlign~\cite{shen2017style} disentangles the style of text and contents via shared latent space for style revision. (ii) StyleEmbeddedc~\cite{fu2018style} and (iii) MultiDec~\cite{fu2018style} extract out style information from text and encode it into embeddings and decoders respectively. (iv) UnsupMT~\cite{zhang2018style} adopts machine translation methods to deliver pseudo training pairs for sequence-to-sequence transduction. (v) \masker~\cite{malmi2020padlm}, a recently proposed unsupervised method for text style transfer, is closest to \oreo. It employs a BERT which masks the span according to the disagreement of language models conditioned on different attributes and fills in a new span for the target attribute. For a fair comparison, we use \robert as their base model. In our preliminary experiment, we find that \robert leads to better performance on text formalization.

\section{Experiment Results}

\subsection{Automatic Evaluation}\label{sec:auto_eval}

\paragraph{Text simplification} Table~\ref{tab:simp_full} presents the automatic evaluation results for text simplification on \newturk. As for the main metric of text simplification, our method achieves the highest SARI score, surpassing the supervised and unsupervised baseline by a large margin. According to \cite{maddela2020controllable}, \texttt{Add} is an important metric to indicate the model's capability in paraphrasing. \oreo gains a higher \texttt{Add} score than the supervised edit-based method, EditNTS. Although UNTS is on a par with \oreo in FKGL scores, its \texttt{Add} score is 0.8 points lower than \oreo, indicating that our model has a better trade-off between simplicity and meaning preservation as well as fluency. Our method's high score in \texttt{Keep} and \texttt{Delete} operations demonstrates that gradient-guided span selection can detect the complex span accurately.

\paragraph{Text formalization}
Table~\ref{tab:style_full} shows the evaluation results for text formalization. 
Our approach outperforms all of the unsupervised baseline models in both content preservation and accuracy of style transfer. Notably, the significant margin of \oreo and \masker demonstrates the necessity of hidden states optimization. Although both methods directly conduct span replacement, \oreo additionally performs on-the-fly update on hidden representations of its context, which is steered by an attribute head. This leads to a large improvement in formality. Additionally, \masker proposes phrasal replacement based on an incomplete input, without accessing the semantics of the original span. This leads to semantic loss. While our span infilling is conditioned on the representations encoded the semantics of the original input, \oreo has a large improvement on BLEU score.

\subsection{Human Evaluation}
To verify the improvement of \oreo, we conduct human evaluation on text formalization in Table~\ref{tab:style_human}. We randomly sample 80 examples from each model's output and human-written reference. Due to the budget limits, we only compare to the baseline that is closest to our work. We invited six annotators with advanced linguistic backgrounds to evaluate formality, semantic coherence and language fluency of each sentence in a blind manner. 
Formality indicates to how much degree the output satisfies the formal attribute. Semantic coherence means whether the output preserves the original semantics of input text. And language fluency measures the grammatical correctness of the output text.
Each annotator is asked to provide scores from 1 to 4 for all three criteria. Each sentence is rated by two annotators~\footnote{The annotators' ratings are positively correlated with $p$-value $<$ 0.1 across models and metrics.} and we report the averaged ratings.  In Table~\ref{tab:style_human}, \oreo is significantly better than \masker in terms of formality and coherency ($p$-value $<$ 0.01), which is consistent with automatic evaluation results. However, there is still improvement space for \oreo when compared to human reference. Two edit-based methods have the same score of language fluency, mostly because both of them recruit \robert as the base model to propose new span.

\begin{table}
        \centering
        \small
        \resizebox{0.8\linewidth}{!}{
        \begin{tabular}{c|c|c|c}
            		\toprule
            		 & Formality  & Coherency & Fluency \\ \midrule
            		 \masker  
            		& 2.74 & 2.94 & 3.31   \\ 
            		\oreo  
            		& 3.42 & 3.33 & 3.41 \\  
            	    Human
            		& \textbf{3.69} & \textbf{3.67} & \textbf{3.78}  \\
            		\bottomrule
        \end{tabular}
           }
        \caption{\small Human evaluation on text formalization}
        \label{tab:style_human}
\end{table}

\begin{table}[!t]
        \centering
        \resizebox{\linewidth}{!}{
            \begin{tabular}{l|c|c|cc}
        		\toprule
        		 & BLEU  & Formality & H-mean & G-mean \\ \midrule
        		 Full
        		& \textbf{57.63} & \textbf{80.71} & \textbf{67.24} & \textbf{68.20} \\  \midrule
        		(1) Infill w/o $H^{(i)}$
        		& 55.50 & 69.67 & 61.78 & 62.18 \\
        		(2) Update $H^{(i)}$ w/ noise
        		& 56.55 & 69.14 & 62.21 & 62.53  \\
        		(3) Fix $H^{(i)}$ 
        		& 56.47 & 67.94 & 61.68 & 61.94  \\
        		(4) Random span selection
        		& 45.30 & 55.03 & 49.69 & 49.93  \\
        		\bottomrule
        	\end{tabular}
            }
        \caption{Model ablation study on text formalization.}
        \label{tab:style_abl}
\end{table}

\subsection{Analysis}\label{subsec:analysis}

\paragraph{Ablation study} We evaluate different variants of \oreo in Table~\ref{tab:style_abl}. To verify the necessity of infilling conditioned on updated hidden states and the gradient information for the update, we compare to variants as 1) without fixing any hidden state when infilling span; 2) updating the hidden states with Gaussian noise; 3) without updating the hidden states. To evaluate the effect of our span selection strategy, we also try (4) randomly selecting span.

With fixed or incorrectly updated hidden states, the formality of revised text drops sharply. It indicates that optimizing hidden states efficiently is crucial to infilling a span that satisfies the target attribute.

When the hidden states are removed, there is a significant drop in terms of the BLEU score due to the loss of semantic information. Both BLEU score and formality drop drastically when the span is replaced randomly. It indicates that our gradient-guided span selection is helpful in detecting spans that are opposite to the target attribute.

\paragraph{Case study}  Table~\ref{tab:case} exhibits the examples generated by baseline methods and \oreo in both tasks. Compared to other baseline methods, our \oreo is able to produce accurate and fluent revision. More surprisingly, it can even conduct knowledgeable revision. For instance, ``\textit{a think tank}'' is simplified as ``\textit{a group that studies people}''. 
\oreo also has decent performance encountering noisy text. In Example 4, \masker fails to correct the abbreviation and typos, while \oreo correctly revises ``\textit{u}'' to ``\textit{you}'', and ``\textit{kno}'' to ``\textit{know}''.

However, we also notice that \oreo sometimes fails to hold semantics. For instance, it revises ``\textit{critics}'' to ``\textit{supporters}'' in Example 4. This is a common problem that language models are not sensitive to negation. More efforts could be made in future work.

\begin{table*}[t!]
    \small
    \centering
    \resizebox{\textwidth}{!}{
    \begin{tabular}{p{1cm}p{6cm}p{6cm}p{6cm}}
        \toprule
        \!\#\! & \textbf{Complex Input} & \textbf{UNTS} & \textbf{\oreo}
        \\ \midrule
        \!\!1\! 
        & kraft announced monday that it will remove artificial food coloring, notably yellow no. 5 and yellow no. 6 dyes, from its iconic product by january 2016.
        & announced \textcolor{red}{announced tuesday} that it will remove artificial food color, including yellow yellow, no yellow, yellow no simple, meaning, from its \textcolor{blue}{famous} product by \textcolor{red}{september}.	
        & kraft announced monday that it will \textcolor{blue}{stop using some of the chemicals, such as} yellow no. 5 and yellow no. 6 dyes, from its iconic product by january 2016. 
        \\ \midrule
        \!\!2\!
        & still, recent trends suggest seattle is doing a better job of holding onto those kids, according to sightline institute, a think tank based in seattle.
        & still, recent trend suggest seattle is doing a better job of holding \textcolor{red}{guns} of those kids, according to \textcolor{red}{unc}, a think tank in seattle.
        & still, recent studies suggest seattle is doing a better job of holding onto those kids, according to sightline institute, a \textcolor{blue}{group that studies people} in seattle.\\ \midrule
        \!\!3\!
        & critics of the program say the eisenhower deportation program's conditions were anything but humane.	
        & critics of the program say the \textcolor{red}{nsa operation}  program's \textcolor{red}{conditions's} conditions were anything.
        & some \textcolor{red}{supporters} of the program say the eisenhower school program's \textcolor{blue}{rules} were anything but for children.
        \\ \midrule \midrule

        \!\#\! & \textbf{Informal Input} & \textbf{\masker} & \textbf{\oreo}\\
        \midrule
         \!\!4\!
         & tell him, and it wouldn't seem psycho cuz u have kno each other for a long time 
         & It wouldn't seem psycho cuz u have kno each other for a long time 
         & Tell him, and it \textcolor{blue}{will not} even seem \textcolor{blue}{awkward you two} have \textcolor{blue}{known} each other for a long time
         \\ \midrule
         \!\!5\!
         & As a guy - let me say this. 
         & As a \textcolor{red}{Christian}\textcolor{blue}{, } I will say this. 
         & As a guy\textcolor{blue}{, please }let me say this. 
         \\ \midrule
         \!\!6\!
         & Intellect - a chick with brains is just sexy!
         & Intellect - is just sexy!
         & \textcolor{blue}{I think a woman endowed} with brains is just sexy! 
         \\ \midrule
        \!\!7\!
        & They bother U all day long. 
        & They \textcolor{red}{are about a long}. 
        & They bother \textcolor{red}{me} \textcolor{blue}{constantly} all day long. 
        \\ \bottomrule
    \end{tabular}
    }
    \caption{Examples of outputs from baseline methods and \oreo on text simplification and text formalization. Both successful and erroneous cases are reported. \textcolor{red}{Red} marks the wrong revision; \textcolor{blue}{Blue} marks the good revision.}
    \label{tab:case}
\end{table*}

Then we explore human-in-the-loop generation, where a user selects a phrase to be replaced; based on which \oreo conducts the revision. We find that this interactive generation can help \oreo conduct better revision.
Examples are in Table 6 in the Appendix B.

\paragraph{Inference efficiency} An obvious concern of \oreo is the inference efficiency, given that it updates the hidden states in a large Transformer on the fly and conducts revision in multiple iterations. Therefore, we report the inference speed here. For text formalization, it takes an average of 0.12 second to revise a sentence in one iteration in \oreo and 4.18 seconds in \masker. We argue that this is acceptable given training in \oreo is simple and time-saving. Moreover, to further reduce the inference duration, we can employ \oreo to construct pseudo-parallel datasets, and learn a conventional sequence generation model as in~\citet{malmi2020padlm}.

\section{Related Work}

\paragraph{Unsupervised text generation} Neural text generation with non-parallel data has received great attention. One approach is defining a pre-defined reward function to guide the training of policy for text generation~\cite{siddique2020dl4pps}. Another one is based on variational auto-encoders, transferring the attributes, such as sentiment~\cite{hu2017sentitrans}, syntax~\cite{chen2019disentangle}, and toxicity~\cite{dos2018detoxic}, by modeling and manipulating the latent variables. 

In this work, we consider the approaches with much simpler training methods. Recently, an approach based on iterative local edit for text revision has been developed. This approach sets an objective function, randomly proposes a set of candidates, and employs discrete optimization algorithms, such as Metropolis–Hastings sampling~\cite{miao2019cgmh} and simulated annealing~\cite{liu2020sa, Li2020UnsupervisedTG}, to accept or reject proposed candidates. Though the training of this approach is simple, the inference is computationally expensive. It has to evaluate a large set of randomly proposed candidates and train multiple neural models for evaluation. 
Our \oreo, however, is much more efficient thanks to the optimized hidden states when revising text.

\paragraph{Steering pre-trained models for text generation}
Our work is also closely related to a brand-new line of research, steering a pre-trained language model to control text generation. Multiple methods of steering have been proposed, one of which is steered by prompt.~\citet{wallace2019universaltrig} finds a universal prompt to trigger a GPT-2 model to generate toxic content.~\citet{chan2020cocon} incorporates content-conditioner block into the GPT-2 model to do a fine-grained control of the attribute for open-domain text generation. 

In this work, we adopt a different approach, steering the hidden states of the pre-trained Transformer. Plug-and-play language model~\cite{dathathri2019plug} is related to our \oreo in the sense that it also updates the hidden states during inference. We highlight the difference between them in two aspects. First, they tackle the task of open-domain text generation, while we consider text revision, which has a constraint from the source (input) text. And hence, we have different generation methods (our iterative span replacement v.s. their conventional left-to-right decoding) and choices of base model (our bi-directional \robert v.s. their unidirectional GPT-2). Second, the steering of hidden states is different. While they employ an additional plug-in module, we let \robert update according to its own estimation.

\paragraph{Text simplification}
Most of the existing work on text simplification relies on the parallel corpus. For instance,~\citet{zhang2017rl4nts} casts simplification into the framework of reinforcement learning.~\citet{dong2019editnts} suggests explicitly modeling the edit operations.~\citet{maddela2020controllable} proposes a pipeline, where the first part focuses on syntactic simplification, while the second part focuses on lexical and phrasal simplification. Recently, there have been efforts made for unsupervised text simplification.~\citet{surya2018unts} employs the idea of variational auto-encoder.~\citet{kumar2020unsupedit4nts} parses the sentence to a constituency tree, conditioned on which they conduct syntactic simplification. None of those work optimizes the distributed representation of text.

\paragraph{Text style transfer}
Variational auto-encoder (VAE) and adversarial learning~\cite{shen2017style, hu2017toward, fu2018style} are well-adopted ideas for text style transfer, which aims to disentangle the style and content of texts in latent space. Due to the issue of computational inefficiency and unstable training, some simpler approaches propose to edit partial texts of input. 
\citet{li2018delete} replaces the stylized n-grams with retrieved alternative words with target style. 
\citet{reid2021lewis} constructs pseudo parallel corpus to train a tagger model and predict token-level edit operations to guide revision. 
\citet{malmi2020padlm} is relatively close to \oreo in the way that it conducts in-place span replacement for style transfer. However, their replacement is not conditioned on on-the-fly optimized hidden states, which has been found in our experiments to be critical for transferring the attribute and preserving semantics. And we use a totally different span selection method.

\section{Conclusion}
In this paper, we propose a new method for text revision with iterative in-place span replacement. With simple fine-tuning methods, the hidden states of \robert can be optimized towards the target attribute on the fly. Both the automatic evaluation and the human evaluation demonstrate the effectiveness of the proposed method in real-world applications, text simplification and text formalization. In the future, we would like to apply this method to more challenging attributes, e.g. modifying syntax for paraphrasing~\cite{chen2019disentangle} and question generation~\cite{Li2019ImprovingQG, gao2020discern}.

\section*{Acknowledgements}
The work described in this paper was supported by the National Key Research and Development Program of China (No. 2018AAA0100204) and Research Grants Council of the Hong Kong Special Administrative Region, China (No. CUHK 14210920 of the General Research Fund). We would like to thank the anonymous reviewers for their comments. 

\bibliography{aaai22}
\newpage

\appendix

\section{Details of Multi-task Fine-tuning}\label{app:detail_ft}
\paragraph{Masked languagde modeling} For the standard MLM objective, we replace 15\% tokens as $\texttt{[MASK]}$. \\
For the padded variant of MLM, we replace one text span with 3 $\texttt{[LM-MASK]}$s for each training instance. If the length of selected span is less than 3, we append $\texttt{[PAD]}$ tokens to it as the target of padded MLM. For example, we mask the first two words in sentence ``Good luck to you!'' as ``$\texttt{[LM-MASK]} \texttt{[LM-MASK]} \texttt{[LM-MASK]}$ to you!'', and the target is ``Good luck $\texttt{[PAD]}$ ''. \\
In this way, we construct two separate sets of MLM training data.

\section{Human-in-the-loop Generation}\label{app:human-loop}
We further consider human-in-the-loop generation, where we let a user decide the phrase to be edited, and use \oreo to make prediction. The examples are demonstrated in Table~\ref{tab:human-loop}. When a user specifies an accurate span, \oreo can correctly revise the sentence to a former tone. For example, \oreo failed to automatically detect the unmatched parentheses in Example 5, but it correctly edits the random punctuation after the user points it out. Additionally, we also select some imperfect outputs from \oreo. Based on its own system outputs, we can obtain better rewrites by explicitly providing spans to be edited.

\begin{table*}[t!]
    \centering
    \resizebox{0.8\textwidth}{!}{
        \begin{tabular}{l}
            \toprule
            \textbf{informal}: \underline{I'm just} looking for the girl who wants that time with me \\
            \textbf{\oreo}: I am not just looking for the girl who wants that time with me \\
            \textbf{1$_{st}$ Edit}: \textcolor{orange}{I am still} looking for the girl who wants \underline{that} time with me \\
            \textbf{2$_{nd}$ Edit}: I am still looking for the girl who wants \textcolor{orange}{to spend} time with me \\ 
            \midrule
            \textbf{informal}: Then if he still \underline{doesn't} like you, then move on.\\
            \textbf{\oreo}: Then if he still does not not like you, then move on. \\
            \textbf{1$_{st}$ Edit}: Then if he still \underline{\textcolor{orange}{does not not}} like you, then move on.\\
            \textbf{2$_{nd}$ Edit}: \underline{Then} if he still \textcolor{orange}{does not appear to} like you, then move on. \\
            \textbf{3$_{rd}$ Edit}: \textcolor{orange}{However,} if he still does not appear to like you, \underline{then} move on. \\
            \textbf{4$_{th}$ Edit}: However, if he still does not appear to like you, \textcolor{orange}{you should} move on. \\
            \midrule
            \textbf{informal}: Maybe your young, I \underline{don't} know. \\
            \textbf{\oreo}: Maybe your young, I do not really know. \\
            \textbf{1$_{st}$ Edit}: Maybe \underline{your} young, I \textcolor{orange}{do not} really know. \\
            \textbf{2$_{nd}$ Edit}: Maybe \textcolor{orange}{you are} young, I do not \underline{really know}. \\
            \textbf{3$_{rd}$ Edit}: Maybe you are young, I do not \textcolor{orange}{know your age}. \\
            \midrule
            \textbf{informal}: No matter what he says, Action speaks louder \underline{then} words \\
            \textbf{\oreo}: No matter what he says, be patient! \\
            \textbf{1$_{st}$ Edit}: No matter what he says, action speaks louder \textcolor{orange}{than verbal} words \\
            \midrule
            \textbf{informal}: ( even if he  is rich, \underline{dont} sell ur self. \\
            \textbf{\oreo}: (even if he is rich, be your own self. \\
            \textbf{1$_{st}$ Edit}: \underline{(even} if he is rich, \textcolor{orange}{do not} sell ur self. \\
            \textbf{2$_{nd}$ Edit}: \textcolor{orange}{even} if he is rich, do not sell \underline{ur self}. \\
            \textbf{3$_{rd}$ Edit}: even if he is rich, do not sell \textcolor{orange}{yourself to him}. \\
            \midrule
            \textbf{informal}: I \underline{haven't} tried it yet, but I am still trying to get up the nerve. \\
            \textbf{\oreo}: I have not even tried it yet, but I am still trying to get up the nerve. \\
            \textbf{1$_{st}$ Edit}:I \textcolor{orange}{have not} even tried it yet, but I am still trying to get up the \underline{nerve}. \\
            \textbf{2$_{nd}$ Edit}: I have not even tried it yet, but I am still trying to \underline{get up the} \textcolor{orange}{courage}.\\
            \textbf{3$_{rd}$ Edit}:  I have not even tried it yet, but I am still trying to \textcolor{orange}{find the \underline{strength} \underline{courage}}.\\
            \textbf{4$_{th}$ Edit}: I have not even tried it yet, but I am still trying to find the \textcolor{orange}{courage}. \\
            \midrule
            \midrule
            \textbf{informal}: Then see if shes open for a dinner \& a movie. \\
            \textbf{\oreo}: \underline{Then see} if she will accompany you for a dinner or perhaps a movie. \\
            \textbf{1$_{st}$ Edit}: \textcolor{orange}{Inquire her} if she will accompany you for a dinner or perhaps a movie.\\
            \midrule
            \textbf{informal}: The same guy you wanna be in a relationship with? \\
            \textbf{\oreo}: \underline{The} same guy you want to be in a relationship with? \\
            \textbf{1$_{st}$ Edit}: \underline{\textcolor{orange}{The very}} same guy you want to be in a relationship with? \\
            \textbf{2$_{nd}$ Edit}: \textcolor{orange}{Is this the} same \underline{guy} you want to be in a relationship with? \\
            \textbf{3$_{rd}$ Edit}: Is this the same \textcolor{orange}{person} that you want to be in a relationship with? \\
            \midrule
            
            \textbf{informal}: if he really didn't like her and did like you, then he would have already dumped her for you. \\
            \textbf{\oreo}: If he really did not like her and did like you, then he would have already \underline{dumped} her for you.\\
            \textbf{1$_{st}$ Edit}: if he really did not like her and did like you, then he would have already \textcolor{orange}{left} her for you. \\

            \bottomrule
        \end{tabular}
        }
    \caption{Examples of human-in-the-loop. Input sentences are edited in multiple iterations. The underlined \underline{texts} are selected span-to-edit. \textcolor{orange}{Orange} indicates proposed phrasal replacement.}
    \label{tab:human-loop}
\end{table*}

\end{document}